\documentclass[11pt]{article}

\usepackage[preprint]{acl}

\usepackage{times}
\usepackage{latexsym}

\usepackage[T1]{fontenc}

\usepackage[utf8]{inputenc}

\usepackage{microtype}

\usepackage{inconsolata}

\usepackage{graphicx}

\usepackage{amsfonts}
\usepackage{amsmath}
\usepackage{amssymb}

\usepackage{booktabs}
\usepackage{multirow} 
\title{SGR: A Stepwise Reasoning Framework for LLMs with External Subgraph Generation}

\author{Xin Zhang\textsuperscript{1}, Yang Cao\textsuperscript{2}, Baoxing Wu\textsuperscript{1}, Kai Song\textsuperscript{2}, Siying Li\textsuperscript{1} \\
$^1$School of Information Science and Engineering, Chongqing Jiaotong University\\ $^2$School of Computer Science and Technology, Chongqing University of Posts and Telecommunications\\
}

\begin{document}
\maketitle
\begin{abstract}
Large Language Models (LLMs) have demonstrated strong capabilities across diverse NLP applications, such as translation, text generation, and question answering. Nevertheless, they remain limited in complex settings that demand deep reasoning and logical inference. Since these models are trained on large-scale text corpora, their generation process may still introduce irrelevant, noisy, or factually inconsistent content.
To mitigate this problem, we introduce SGR, a stepwise framework that enhances LLM reasoning through external subgraph generation. SGR builds query-specific subgraphs from external knowledge bases and uses their semantic structure to support multi-step inference. By grounding intermediate reasoning steps in structured external knowledge, the framework helps the model concentrate on relevant entities, relations, and supporting evidence.
In particular, SGR first constructs a subgraph tailored to the input question. It then guides the model to reason progressively over the generated structure and combines multiple reasoning trajectories to obtain the final prediction. Experimental results across several benchmark datasets show that SGR achieves consistent improvements over competitive baselines, highlighting its value for improving both reasoning accuracy and factual reliability.
\end{abstract}

\section{Introduction}

Recent advances in large language models (LLMs) have become a major milestone in the development of artificial intelligence. By pretraining on large-scale textual corpora, these models can produce fluent and high-quality text \citep{brown2020language}. With additional supervised fine-tuning, they can also be adapted to various downstream natural language processing tasks \citep{wei2022chain}. Nevertheless, despite their impressive generation ability, LLMs still exhibit notable weaknesses in complex reasoning settings. In particular, while they are effective at generating coherent text, they may fail to preserve logical consistency, resulting in outputs that conflict with real-world facts. In addition, because of their large parameter scale and opaque internal mechanisms, the reasoning behavior of LLMs is often difficult to interpret \citep{jacovi2020towards}.

One promising way to mitigate these issues is to incorporate external knowledge resources, especially knowledge graphs (KGs), into LLM-based reasoning. KGs provide an explicit and structured representation of factual knowledge, which can compensate for some limitations of parametric language models \citep{ji2022survey}. Prior work has investigated the use of KGs as supplementary knowledge sources for improving the reasoning ability of LLMs \citep{lewis2020retrieval,yao2014information}. In general, these approaches introduce KG-derived information into LLMs through prompt construction or additional contextual inputs, thereby supporting the reasoning process. Beyond KG-based enhancement, recent research has also studied the integration of LLMs with symbolic or structural representations to improve systematic generalization and interpretability \citep{yang2024harnessing, yang2024can, xiong2025deliberate}. However, many existing KG-injection methods still depend on fixed or manually crafted prompt templates and therefore fail to sufficiently utilize the rich structural information encoded in knowledge graphs \citep{jiang2023structgpt,li2023tog}.

Motivated by this observation, we introduce SGR, a stepwise reasoning enhancement framework for LLMs based on external subgraph generation. SGR uses schema-guided subgraph construction over knowledge graphs to identify query-relevant knowledge more accurately. By dynamically building external subgraphs, the framework guides LLMs to reason progressively, which helps reduce the impact of irrelevant or noisy information and improves reasoning accuracy. Moreover, SGR combines direct reasoning with integrated reasoning strategies to generate the final enhanced output.

Experiments conducted on multiple benchmark datasets show that SGR outperforms existing methods. The main contributions of this paper are summarized as follows.

\begin{itemize}
    \item We introduce a new stepwise reasoning enhancement framework that strengthens LLM reasoning by incorporating externally generated subgraphs. This method improves the capacity of LLMs to address complex multi-step reasoning problems.
    \item We develop an interpretable knowledge-enhanced reasoning procedure that increases the transparency of LLM reasoning. The generated reasoning paths and final outputs can be traced and verified.
    \item We conduct extensive experiments on several benchmark datasets, confirming the effectiveness of the proposed framework and showing its advantages over prior approaches.
\end{itemize}

\section{Related Work}

Knowledge graphs organize extensive external world knowledge as structured triples and have been widely adopted for knowledge representation and reasoning \citep{nickel2016review}. Early research investigated how structured knowledge could be incorporated into neural models \citep{sun2019bert}. More recently, many studies have attempted to strengthen LLMs by injecting KG-derived knowledge at either the training or inference stage, leading to improved performance on diverse tasks \citep{liu2023survey}. In addition to static KGs, temporal knowledge graphs have been developed to model the evolving states of entities and relations across time, supporting richer reasoning under temporal constraints \citep{xiongtilp,xiong2024teilp, xiong2024large}. Nevertheless, directly encoding KG knowledge into LLM parameters can substantially increase training costs and limit model flexibility \citep{logan2019barack}.

To alleviate this problem, recent work has explored retrieving query-relevant knowledge from KGs at inference time and supplying it to LLMs as supplementary context \citep{lewis2020retrieval,guu2020retrieval}. Existing methods in this direction can generally be divided into retrieval-enhanced approaches and collaborative enhancement approaches.

\subsection{Retrieval Enhanced Models}

Retrieval-enhanced models aim to obtain relevant information from an external knowledge source and feed it into the language model, helping the model generate more accurate answers \citep{lewis2020retrieval}. Jiang et al. introduced StructGPT, a structured reasoning retrieval framework that combines iterative reading and reasoning mechanisms \citep{jiang2023structgpt}. Other related studies further investigate multi-hop retrieval and reasoning over structured knowledge to improve answer accuracy \citep{das2019multi,chen2020open}.

\begin{figure*}[t]
  \centering
  \includegraphics[width=0.8\linewidth]{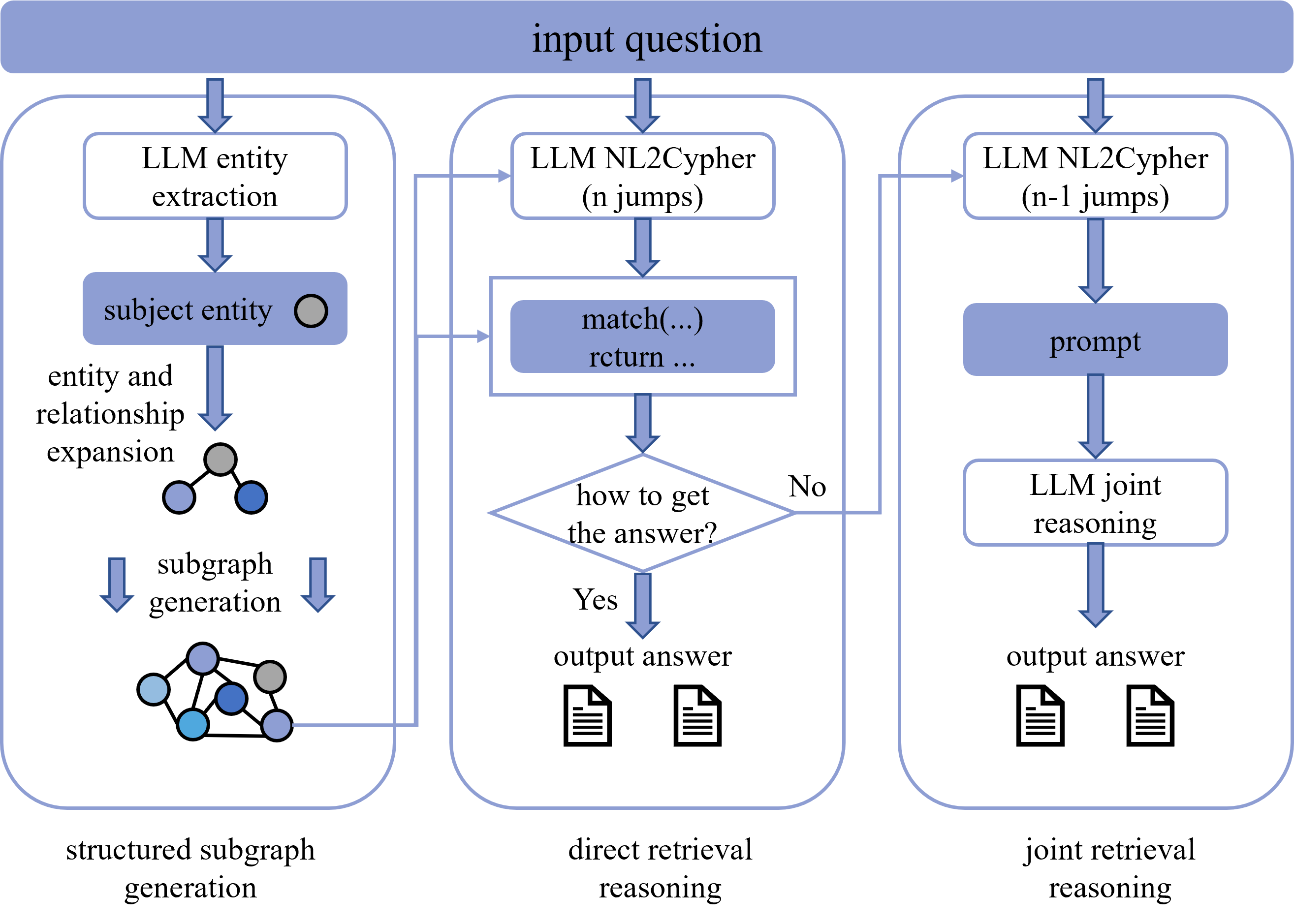}
  \caption{Pipeline of SGR framework.}
  \label{fig:fig1}
\end{figure*}

Despite their effectiveness, retrieval-enhanced approaches still have several practical limitations. First, the retrieved content may fail to cover the complete structured reasoning paths encoded in the knowledge graph, leading to incomplete reasoning chains \citep{das2019multi}. Second, retrieving too much information can introduce irrelevant evidence and overwhelm the model, which may reduce reasoning performance \citep{izacard2021leveraging}. Finally, when the retrieved knowledge includes errors or redundant facts, it can negatively influence the final reasoning outcome \citep{petroni2021kilt}.

\subsection{Collaborative Enhanced Models}

Collaborative enhancement methods focus on building interaction mechanisms between LLMs and knowledge graphs. Under this paradigm, both the language model and the KG participate in the reasoning process to produce final answers \citep{luo2022knowledge,zhou2022least}. Typically, these methods first use the language model to generate an initial reasoning result, then employ the knowledge graph to validate or refine intermediate outputs, and finally obtain an enhanced answer. For example, \citet{luo2022knowledge} proposed a KG-enhanced reasoning framework in which the language model performs initial reasoning over the query, after which the knowledge graph is used for validation and correction. By iteratively interacting with the KG, the system progressively improves reasoning accuracy.

Compared with retrieval-enhanced methods, collaborative enhancement approaches place greater emphasis on bidirectional interaction between the language model and the knowledge graph. In this setting, the language model not only receives graph-based knowledge, but also produces intermediate hypotheses that guide subsequent KG queries \citep{yang2023collaborative}. Although this paradigm can improve reasoning robustness, it continues to face challenges in interaction efficiency and reasoning consistency.

\section{Methodology}

To more effectively combine the reasoning ability of large language models with the structured knowledge provided by external knowledge graphs, we introduce SGR, a stepwise reasoning enhancement framework based on external subgraph generation. As illustrated in the overall framework, SGR consists of three major components: structured subgraph construction, stepwise reasoning enhancement, and collaborative reasoning integration.

Given an input question, the framework first converts it into a structured query and retrieves a relevant subgraph from the knowledge graph. The language model then performs step-by-step reasoning grounded in the generated subgraph to obtain candidate answers. Finally, reasoning results from multiple paths are integrated to produce the final output.

\subsection{Structured Subgraph Generation}

\subsubsection{Knowledge Graph Construction}

Let the external knowledge graph be denoted as
\begin{equation}
\mathcal{G}=(\mathcal{V}, \mathcal{E}, \mathcal{R}),
\end{equation}
where $\mathcal{V}$ is the set of entities, $\mathcal{R}$ is the set of relation types, and $\mathcal{E} \subseteq \mathcal{V} \times \mathcal{R} \times \mathcal{V}$ represents the set of factual triples. Each triple $(h,r,t) \in \mathcal{E}$ indicates that a head entity $h$ is connected to a tail entity $t$ through relation $r$.

For an input question $q$, SGR first constructs a structured representation by extracting key entities, relations, and constraints:
\begin{equation}
\mathcal{S}_q = \{ \mathcal{V}_q, \mathcal{R}_q, \mathcal{C}_q \},
\end{equation}
where $\mathcal{V}_q \subseteq \mathcal{V}$ denotes the question-related entities, $\mathcal{R}_q \subseteq \mathcal{R}$ denotes the relevant relations, and $\mathcal{C}_q$ represents additional semantic or logical constraints derived from the question.

The objective of subgraph construction is to identify a compact query-relevant subgraph
\begin{equation}
\mathcal{G}_q = (\mathcal{V}_q^\ast, \mathcal{E}_q^\ast, \mathcal{R}_q^\ast),
\end{equation}
where $\mathcal{V}_q^\ast \subseteq \mathcal{V}$, $\mathcal{R}_q^\ast \subseteq \mathcal{R}$, and $\mathcal{E}_q^\ast \subseteq \mathcal{E}$. This subgraph serves as the structured knowledge foundation for later reasoning.

The generated subgraph provides explicit relational paths that guide the language model during reasoning. Compared with unstructured textual context, this structured representation allows finer control over the reasoning process and helps reduce the influence of irrelevant information.

\subsubsection{Subgraph Generation Process}

The subgraph construction procedure contains three main steps. First, the language model identifies the central entities and relations in the input question. Second, the knowledge graph is queried to obtain related entities and relations, forming candidate subgraphs. Third, nodes and edges with low semantic relevance to the question are filtered out.

Formally, for each candidate triple $e_i=(h_i,r_i,t_i)$, we compute a relevance score with respect to the input question:
\begin{equation}
s(e_i,q)=\mathrm{sim}\left(\phi(e_i), \phi(q)\right),
\end{equation}
where $\phi(\cdot)$ denotes a semantic encoder and $\mathrm{sim}(\cdot,\cdot)$ measures semantic similarity. The final subgraph is selected by retaining triples whose relevance scores exceed a threshold $\tau$:
\begin{equation}
\mathcal{E}_q^\ast = \{ e_i \in \mathcal{E} \mid s(e_i,q) \geq \tau \}.
\end{equation}

Alternatively, the subgraph can be obtained by selecting the top-$k$ most relevant triples:
\begin{equation}
\mathcal{E}_q^\ast = \operatorname{TopK}_{e_i \in \mathcal{E}} \; s(e_i,q).
\end{equation}

By producing a query-specific subgraph for each input, SGR avoids the restrictions of fixed prompt templates and makes better use of the structural information contained in the knowledge graph.

\subsection{Stepwise Reasoning Enhancement}

Using the generated subgraph, SGR guides the language model to reason in a stepwise manner. Let a reasoning path be represented as
\begin{equation}
p = (e_1, e_2, \ldots, e_T),
\end{equation}
where each $e_t \in \mathcal{E}_q^\ast$ is a selected triple at reasoning step $t$. At each step, the language model generates an intermediate reasoning state $z_t$ conditioned on the question, the previous reasoning states, and the current subgraph evidence:
\begin{equation}
z_t = f_{\theta}(q, z_{<t}, e_t),
\end{equation}
where $f_{\theta}$ denotes the LLM parameterized by $\theta$, and $z_{<t}$ represents previous reasoning states.

The probability of generating a complete reasoning trajectory can be written as
\begin{equation}
P_{\theta}(z_{1:T} \mid q, \mathcal{G}_q)
=
\prod_{t=1}^{T}
P_{\theta}(z_t \mid q, z_{<t}, \mathcal{G}_q).
\end{equation}

To encourage the reasoning process to follow the generated subgraph, we introduce a path consistency score:
\begin{equation}
C(p,\mathcal{G}_q)=\frac{1}{T}\sum_{t=1}^{T}\mathbb{I}(e_t \in \mathcal{E}_q^\ast),
\end{equation}
where $\mathbb{I}(\cdot)$ is an indicator function. A higher score indicates that the reasoning path is more consistent with the retrieved structured knowledge.

The final stepwise reasoning objective can be described as selecting the answer with the highest joint reasoning confidence and structural consistency:
\begin{equation}
\hat{a}
=
\arg\max_{a}
P_{\theta}(a \mid q, \mathcal{G}_q)
\cdot
C(p,\mathcal{G}_q).
\end{equation}

This step-by-step reasoning mechanism allows the model to decompose complex reasoning problems into smaller subproblems. By explicitly following relational paths in the generated subgraph, the model can reduce logical inconsistency and improve reasoning accuracy.

\subsection{Collaborative Reasoning Integration}

In the final stage, SGR combines multiple reasoning paths obtained from different traversals of the subgraph. Suppose the framework generates $M$ candidate reasoning paths:
\begin{equation}
\mathcal{P}=\{p_1,p_2,\ldots,p_M\}.
\end{equation}
Each path $p_i$ produces a candidate answer $a_i$ with confidence score $\alpha_i$. The confidence of each path is computed based on both language model probability and graph consistency:
\begin{equation}
\alpha_i =
\lambda P_{\theta}(a_i \mid q, p_i)
+
(1-\lambda) C(p_i,\mathcal{G}_q),
\end{equation}
where $\lambda \in [0,1]$ controls the balance between model confidence and structural consistency.

The final answer is obtained by aggregating candidate answers from different reasoning paths:
\begin{equation}
\hat{a}
=
\arg\max_{a}
\sum_{i=1}^{M}
\alpha_i \cdot \mathbb{I}(a_i=a).
\end{equation}

This integration step improves robustness by reducing the influence of errors from any single reasoning path. Overall, SGR provides a unified framework that brings together structured knowledge, stepwise inference, and collaborative integration to strengthen the reasoning ability of large language models.

The model first conducts entity-based reasoning to generate a structured subgraph schema. This schema represents the external knowledge relevant to the input query. Specifically, the language model is prompted to act as a knowledge graph expert and is provided with several knowledge graph triples. It is asked to reason over these triples and construct a structured query schema.
The model first extracts core entities and relations from the question, and then organizes them into a schema that captures the logical structure required for reasoning. For example, given a question, the model may identify a chain of entities and relations, such as entity one linked by a relation to entity two, entity two connected to entity three, and entity three further connected to entity four. The resulting schema consists of entity-relation pairs that describe the reasoning path needed to answer the question.
During this procedure, SGR uses schema-based reasoning to narrow the search space for knowledge graph queries. The generated schema is then used to construct a structured query in Neo4j, enabling accurate retrieval of relevant knowledge.

\begin{table*}[t]
\centering
\small
\caption{Performance comparison of different reasoning methods on CWQ, WebQSP, and GrailQA. 
Hits@1 and accuracy are reported where applicable, and the best results for each metric are highlighted in bold.
Note: Best results are taken from prior work, including $\alpha$,
$\beta$,
$\gamma$,
and $\delta$.}
\label{tab:experimental_results}
\begin{tabular}{lcccccc}
\toprule
\multirow{2}{*}{Method} &
\multicolumn{2}{c}{CWQ} &
\multicolumn{2}{c}{WebQSP} &
\multicolumn{2}{c}{GrailQA} \\
\cmidrule(lr){2-3} \cmidrule(lr){4-5} \cmidrule(lr){6-7}
& Hits@1 & Acc & Hits@1 & Acc & Hits@1 & Acc \\
\midrule
IO Prompt / ChatGPT        & 0.376 & 0.256 & 0.633 & 0.582 & 0.294 & 0.223 \\
CoT / ChatGPT              & 0.388 & 0.258 & 0.622 & 0.577 & 0.281 & 0.201 \\
\midrule
Prior FT SOTA              & 0.704$^{\alpha}$ & -- & 0.821$^{\beta}$ & -- & 0.754$^{\gamma}$ & -- \\
Prior Prompting SOTA       & -- & -- & 0.744$^{\delta}$ & -- & 0.532$^{\delta}$ & -- \\
\midrule
SGR / Cypher LLM         & 0.523 & 0.445 & 0.745 & 0.706 & 0.624 & 0.593 \\
SGR / ChatGPT            & 0.578 & 0.526 & 0.801 & 0.784 & 0.713 & 0.633 \\
StructGPT / ChatGPT        & -- & -- & 0.726 & -- & -- & -- \\
ToG / ChatGPT              & 0.571 & -- & 0.762 & -- & 0.687 & -- \\
ToG / GPT-4                & \textbf{0.725} & -- & 0.826 & -- & \textbf{0.814} & -- \\
SGR / GPT-4              & 0.632 & \textbf{0.590} & \textbf{0.826} & \textbf{0.808} & 0.756 & \textbf{0.703} \\
\bottomrule
\end{tabular}

\end{table*}

\subsection{Direct Reasoning Enhancement}

\subsubsection{Cypher LLM}

During the direct reasoning enhancement stage, the framework first constructs a Cypher query according to the generated schema. Let the schema produced from the input question $q$ be denoted as
\begin{equation}
\mathcal{S}_q = \{(v_i, r_i, v_j)\}_{i=1}^{L},
\end{equation}
where $v_i$ and $v_j$ represent entities or entity variables, $r_i$ denotes a relation type, and $L$ is the number of schema-level relational constraints.

The goal of the Cypher generation module is to transform $\mathcal{S}_q$ into an executable Cypher query:
\begin{equation}
c_q = g_{\phi}(q, \mathcal{S}_q),
\end{equation}
where $g_{\phi}$ denotes the NLCypher module parameterized by $\phi$, and $c_q$ is the generated Cypher query. NLCypher serves as the language interface that maps natural language questions and structured schemas into Cypher queries executable on the Neo4j database.

The generated Cypher query is then executed over the knowledge graph $\mathcal{G}$ to retrieve candidate answers:
\begin{equation}
\mathcal{A}_q = \operatorname{Exec}(c_q, \mathcal{G}),
\end{equation}
where $\operatorname{Exec}(\cdot)$ denotes the database execution function and $\mathcal{A}_q=\{a_1,a_2,\ldots,a_N\}$ is the set of retrieved candidate answers.

The retrieved results are combined with the original question and provided to the large language model as structured evidence:
\begin{equation}
\hat{a}
=
\arg\max_{a \in \mathcal{A}_q}
P_{\theta}(a \mid q, \mathcal{A}_q, \mathcal{S}_q),
\end{equation}
where $P_{\theta}$ represents the answer probability estimated by the LLM. This formulation allows the model to generate answers grounded in both the natural language question and the retrieved structured knowledge.

In our experiments, we use approximately 140K knowledge graph queries across several datasets. We evaluate Cypher-based reasoning with GPT-based models. The results show that the Cypher-guided reasoning procedure improves question-answering accuracy, especially for complex multi-hop reasoning cases.

\subsubsection{Answer Validation}

After candidate answers are retrieved through Cypher queries, the framework validates them by checking whether they are consistent with the retrieved knowledge and the generated schema. Given a candidate answer $a_i \in \mathcal{A}_q$, we define a validation function:
\begin{equation}
V(a_i, \mathcal{S}_q, \mathcal{G}) =
\mathbb{I}\left[a_i \models \mathcal{S}_q \text{ in } \mathcal{G}\right],
\end{equation}
where $\mathbb{I}[\cdot]$ is the indicator function, and 
$a_i \models_{\mathcal{G}} \mathcal{S}_q$ means that candidate answer $a_i$ satisfies the schema constraints in $\mathcal{G}$.

The valid answer set is therefore defined as
\begin{equation}
\mathcal{A}_q^{+}
=
\{a_i \in \mathcal{A}_q \mid V(a_i, \mathcal{S}_q, \mathcal{G}) = 1\}.
\end{equation}

If $\mathcal{A}_q^{+}$ is non-empty, the final answer is selected from the validated candidate set:
\begin{equation}
\hat{a}
=
\arg\max_{a_i \in \mathcal{A}_q^{+}}
P_{\theta}(a_i \mid q, \mathcal{A}_q^{+}, \mathcal{S}_q).
\end{equation}

Otherwise, if no candidate answer satisfies the schema constraints, the framework triggers additional reasoning steps:
\begin{equation}
\mathcal{A}_q^{+} = \varnothing
\quad \Rightarrow \quad
\text{Refine}(q,\mathcal{S}_q,\mathcal{G}).
\end{equation}

This validation mechanism helps reduce hallucinated responses and maintains consistency between the language model outputs and the underlying knowledge graph.

\subsection{Collaborative Reasoning Enhancement}

When direct reasoning fails to produce a correct answer, the framework applies collaborative reasoning enhancement. In this stage, the language model and the knowledge graph interact iteratively to improve the reasoning process. At iteration $t$, the language model generates an intermediate hypothesis:
\begin{equation}
h_t = f_{\theta}(q, \mathcal{E}_{<t}, h_{<t}),
\end{equation}
where $\mathcal{E}_{<t}$ denotes the evidence retrieved in previous iterations, and $h_{<t}$ represents earlier hypotheses.

The knowledge graph then verifies and expands this hypothesis by retrieving supporting evidence:
\begin{equation}
\mathcal{E}_t = \operatorname{Retrieve}(h_t, \mathcal{G}),
\end{equation}
where $\mathcal{E}_t$ contains the graph triples relevant to the current hypothesis. The retrieved evidence is fed back into the language model for the next reasoning step:
\begin{equation}
h_{t+1}
=
f_{\theta}(q, h_{\leq t}, \mathcal{E}_{\leq t}).
\end{equation}

The iterative process can be expressed as:
\begin{equation}
(q, h_t)
\xrightarrow{\text{KG retrieval}}
\mathcal{E}_t
\xrightarrow{\text{LLM reasoning}}
h_{t+1}.
\end{equation}

After $T$ iterations, the framework obtains a set of candidate reasoning paths:
\begin{equation}
\mathcal{P}=\{p_1,p_2,\ldots,p_M\},
\end{equation}
where each path $p_i$ corresponds to a sequence of hypotheses and retrieved evidence:
\begin{equation}
p_i = (h_1^i, \mathcal{E}_1^i, h_2^i, \mathcal{E}_2^i, \ldots, h_T^i, \mathcal{E}_T^i).
\end{equation}

For each reasoning path, we compute a path score based on model confidence and graph consistency:
\begin{equation}
\beta_i =
\lambda P_{\theta}(a_i \mid q, p_i)
+
(1-\lambda) C(p_i,\mathcal{G}),
\end{equation}
where $\lambda \in [0,1]$ balances language model confidence and structural consistency. The graph consistency term is defined as:
\begin{equation}
C(p_i,\mathcal{G})
=
\frac{1}{T}
\sum_{t=1}^{T}
\mathbb{I}(\mathcal{E}_t^i \subseteq \mathcal{G}),
\end{equation}
where $\mathbb{I}(\cdot)$ is the indicator function.

The final answer is selected by aggregating candidate answers from all reasoning paths:
\begin{equation}
\hat{a}
=
\arg\max_{a}
\sum_{i=1}^{M}
\beta_i \cdot \mathbb{I}(a_i=a).
\end{equation}

Through iterative collaboration, the framework explores multiple reasoning paths and progressively moves toward a more accurate answer. This strategy improves robustness when a single reasoning attempt is insufficient.

\section{Experiments}

\subsection{Experimental Setup}

We evaluate SGR on multiple benchmark datasets, including CWQ, WebQSP, GrailQA, and KQA Pro. CWQ targets complex question answering and requires multi-hop reasoning over knowledge graphs. WebQSP is a commonly used benchmark for Freebase-based question answering. GrailQA focuses on generalization to previously unseen query structures, while KQA Pro covers knowledge-based question answering with diverse reasoning patterns.

For evaluation, we adopt standard metrics, including accuracy and F1 score. Accuracy reflects the percentage of questions answered correctly, whereas F1 measures the balance between precision and recall.

\begin{figure}[t]
  \centering
  \includegraphics[width=0.9\linewidth]{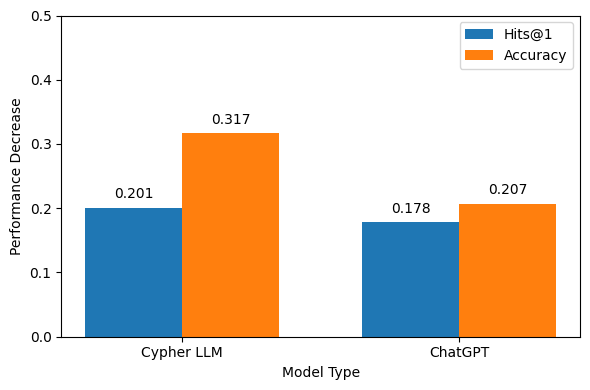}
  \vspace{-10pt}
  \caption{Impact brought by removing Schema prompts.}
  \includegraphics[width=0.9\linewidth]{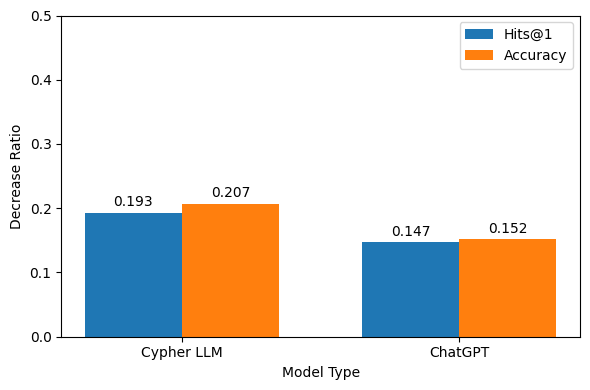}
  \vspace{-10pt}
  \caption{Impact brought by removing neo4j retrieval.}
\end{figure}

\begin{table*}[t]
\centering
\small
\caption{Ablation experiment results on the CWQ dataset}
\label{tab:ablation_cwq}
\begin{tabular}{lcccccccc}
\toprule
\multirow{2}{*}{Method} &
\multicolumn{2}{c}{With Schema} &
\multicolumn{2}{c}{With neo4j} &
\multicolumn{2}{c}{Without Schema} &
\multicolumn{2}{c}{Without neo4j} \\
\cmidrule(lr){2-3} \cmidrule(lr){4-5} \cmidrule(lr){6-7} \cmidrule(lr){8-9}
 & Hits@1 & Acc & Hits@1 & Acc & Hits@1 & Acc & Hits@1 & Acc \\
\midrule
SGR/Cypher LLM & 0.523 & 0.445 & 0.553 & 0.445 & 0.322 & 0.128 & 0.360 & 0.238 \\
SGR/ChatGPT   & 0.578 & 0.526 & 0.578 & 0.526 & 0.400 & 0.319 & 0.431 & 0.374 \\
\bottomrule
\end{tabular}
\end{table*}

\subsection{Experimental Results}

To assess the effectiveness of the proposed method, we conduct extensive experiments on two representative datasets, CWQ and GrailQA, using input-output prompting and chain-of-thought reasoning. We compare SGR with several competitive baselines, including ChatGPT, CoT-based prompting, and previous state-of-the-art methods.

First, we examine the role of external knowledge by evaluating SGR under different settings. The results show that incorporating external subgraph information improves the reasoning performance of large language models. Compared with approaches that depend only on textual prompts, SGR obtains better accuracy and Hits@1 scores on all evaluated datasets, demonstrating the benefit of external subgraph generation.

We also compare SGR against prior state-of-the-art reasoning approaches. The results show that SGR consistently achieves stronger performance on both CWQ and GrailQA. In particular, SGR brings clear improvements even for smaller language models, suggesting that structured external knowledge can help compensate for their limited reasoning capacity.
Furthermore, we study the generalization ability of SGR across different base models. The results indicate that SGR provides consistent gains when applied to both Cypher LLM and ChatGPT, showing that the framework is model-agnostic and can be integrated with various large language models.

\subsection{Ablation Study}

To further examine the contribution of each component, we conduct ablation studies on the CWQ dataset. Specifically, we evaluate the effects of removing schema-guided subgraph generation and Neo4j-based retrieval.

The results show that removing schema-based prompts causes a clear performance drop, confirming the importance of schema guidance for accurate subgraph construction. Similarly, excluding Neo4j-based retrieval substantially decreases reasoning accuracy, indicating that structured knowledge retrieval is crucial for supporting multi-step reasoning.
These findings suggest that both schema guidance and structured retrieval are key components of SGR. Their combination is necessary for achieving the best reasoning performance.

\subsection{Application Scenarios and Error Analysis}

SGR can be used for a broad range of knowledge-intensive reasoning tasks, such as complex question answering and logical inference. In practical settings, SGR shows strong robustness and stability, especially when tasks require multi-step reasoning over structured knowledge.

We further perform error analysis to identify the major causes of remaining failures. The results show that most errors come from incomplete subgraph construction and ambiguous entity linking. In some cases, missing or noisy facts in the external knowledge graph also lead to incorrect reasoning paths. Improving subgraph generation quality and expanding knowledge coverage are therefore important directions for future work.

To further broaden the applicability of SGR, we explore its use in the medical domain by combining medical knowledge graphs with LLMs for complex reasoning tasks, such as disease diagnosis and clinical decision making. By constructing and using structured medical knowledge graphs, SGR helps LLMs reason over medical facts, symptoms, and treatment relations, thereby improving reasoning accuracy and reliability in real-world medical scenarios.
To validate SGR across different domains, we also conduct experiments on academic knowledge graph reasoning tasks. The results show that SGR consistently outperforms baseline methods across multiple datasets. In addition, by dynamically generating external subgraphs, the framework allows LLMs to adapt their reasoning strategies to different contexts, leading to more accurate reasoning and inference.

\section{Conclusion}

This paper presents SGR, a stepwise reasoning enhancement framework for large language models based on external subgraph generation. By dynamically constructing query-relevant subgraphs from knowledge graphs, SGR extracts structured knowledge and guides LLMs through step-by-step reasoning. By combining direct reasoning with collaborative reasoning strategies, the framework improves the accuracy and reliability of reasoning results. Extensive experiments on multiple benchmark datasets demonstrate the strong reasoning enhancement and generalization ability of SGR. Moreover, the framework improves interpretability by providing explicit reasoning paths grounded in external knowledge.

\section*{Limitations}

Although SGR demonstrates strong effectiveness, it still has several limitations. First, the framework incurs extra computational cost because it requires subgraph generation and knowledge retrieval, which may reduce inference efficiency in large-scale applications. Second, its performance is influenced by the completeness and reliability of the underlying knowledge graphs. If the external knowledge contains missing, noisy, or inaccurate information, the reasoning results may be affected. In future work, we will improve the efficiency of subgraph construction and retrieval, and investigate lightweight deployment strategies to make the framework more practical for real-world use.

\bibliography{custom}

\end{document}